\documentclass{llncs}

\usepackage{graphicx}%
\usepackage{multirow}%
\usepackage[title]{appendix}%
\usepackage{xcolor}%
\usepackage{textcomp}%
\usepackage{manyfoot}%
\usepackage{booktabs}%
\usepackage{algorithm}%
\usepackage{algorithmicx}%
\usepackage{algpseudocode}%
\usepackage{listings}%
\usepackage[numbers]{natbib}
\usepackage{times}
\usepackage{latexsym}

\usepackage{qtree}
\usepackage{pict2e}
\usepackage{eepic}
\usepackage{color}
\usepackage{moreverb}
\usepackage{fancyvrb}
\usepackage{tabularx}
\usepackage{colortbl}
\usepackage{acronym}
\usepackage{multirow}
\usepackage{longtable}
\usepackage{paralist}
\usepackage{url}
\usepackage{xspace}
\usepackage{verbatim}			
\usepackage{array}				
\usepackage{colortbl}			
\usepackage{tikz-qtree}
\usepackage{rotating}
\usepackage{etoolbox}
\usepackage{indentfirst}
\usepackage{booktabs}
\usepackage{epstopdf}
\usepackage{bm}

\definecolor{notesColor}{RGB}{134, 179, 0}

\acrodef{NLP}{Natural Language Processing}

\graphicspath{{./figs/}}

\begin{document}

\title{A Study on Bias Detection and Classification in Natural Language Processing}

\author{Ana Sofia Evans\inst{1}\inst{2} \and
Helena Moniz \inst{2}\inst{3}\inst{4} \and
Luisa Coheur \inst{1}\inst{2}}

\authorrunning{A. Evans et al.}
%
\institute{Instituto Superior T\'{e}cnico, Universidade de Lisboa \and INESC-ID \and Unbabel \and Faculdade de Letras, Universidade de Lisboa\\ 
\email{anasevans98@gmail.com} \\ \email{helena.moniz@campus.ul.pt} \\ \email{luisa.coheur@tecnico.ulisboa.pt} }

\maketitle              
%


\abstract{Human biases have been shown to influence the performance of models and algorithms in various fields, including Natural Language Processing. While the study of this phenomenon is garnering focus in recent years, the available resources are still relatively scarce, often focusing on different forms or manifestations of biases. The aim of our work is twofold: 1) gather publicly-available datasets and determine how to better combine them to effectively train models in the task of hate speech detection and classification; 2) analyse the main issues with these datasets, such as scarcity, skewed resources, and reliance on non-persistent data. We discuss these issues in tandem with the development of our experiments, in which we show that the combinations of different datasets greatly impact the models' performance.}

\keywords{Bias Detection, Bias Classification, Hate Speech, Natural Language Processing}


\section{Introduction}\label{intro}

There is a growing awareness of the extent to which human biases can influence our models and algorithms. This realization has led to a fast growth in fields dedicated to studying bias, such as the study of bias in \ac{NLP}, which has focused not only on bias mitigation but also on its detection and classification.  However, bias detection is a relatively new field of study, lacking many publicly available benchmark datasets or state-of-the-art models that are able to complete this task. Existing datasets are relatively small, often do not focus on the same types of bias, and are not even aimed at the same downstream tasks. So, a question arises: \textit{can we learn how to detect and classify bias using these (publicly available) resources? And, if so, how?}

In order to answer this question, we have outlined the following objectives:
\begin{itemize}
    \item Find and collect publicly available datasets aimed at bias classification to serve as training data;
    \item Train and analyse the performance of several classifiers, trained with different parameters and training data combinations;
    \item  Delve into issues such as reliance on non-persistent data, providing both consequence analysis and introducing possible solutions.
\end{itemize}

This work contributes, thus, with a thorough review of the current state-of-the-art in bias and hate speech detection in \ac{NLP}. We provide an overview of various types of work developed in the scope of these fields, complemented by a detailed exposition of some pre-existing datasets which differ in terms of style, collection method, annotation style, and focus. The obtained results allow us to further analyse which aspects of this field require further work, which ones are in desperate need of attention, and possible avenues of future work.

Before we start, we need to define the concept of bias: ``bias" refers to unequal treatment of a given subject due to preconceived notions regarding that very same subject, which necessarily influence our judgement. ``Social bias", therefore, translates to unequal treatment of certain individuals or groups based on specific shared characteristics – namely, social constructs such as race, gender, gender identity, etc. Hence, there are two things that should be defined in order to construct our working definition of bias, namely: what is considered ``unequal treatment"? And what shared characteristics will we consider?

In this paper, we chose to define ``unequal treatment" as:
\begin{itemize}
    \item The use of \emph{derogatory terms} which specifically target an individual or a group based on the defined social characteristics (for example ``bitch", ``dyke", ``tranny");
    \item The prevalence of \emph{stereotypes}, which can also manifest through harmful beliefs (i.e. ``All Muslims are terrorists."), stereotypical societal roles  (i.e. ``Women belong in the kitchen."), caricatures (i.e. ``The Angry Black Woman"), or even apparently benevolent beliefs (i.e. ``Asians are good at math.");
    \item Otherwise abusive language which specifically targets a group or an individual based on the defined social characteristics (i.e. ``Gay people make me sick!", ``I'd never date a black guy.").
\end{itemize}

Additionally, we define that we will be considering the following social characteristics, henceforth referred to as ``Target Categories": Gender, Race, Profession, Religion, Disability, Sexual Orientation, Gender Identity, Nationality, and Age.

In works similar to ours, we find that a term which often approximates our definition of bias is ``hate speech". This is described in \cite{founta2018large}, as ``Language used to express hatred towards a targeted individual or group, or is intended to be derogatory, to humiliate, or to insult the members of the group, on the basis of attributes such as race, religion, ethnic origin, sexual orientation, disability, or gender.” (2018:495) \citep{founta2018large}. Although bias and hate speech share some similarities, they are not quite the same; while instances of Hate Speech will always be instances of bias, the same cannot be said in reverse. However, due to the aforementioned similarities, we will be utilizing resources from both fields.

Considering that the study of bias and hate speech is inherently a sensitive subject, which must be conducted with a degree of awareness and responsibility, we provide an \textbf{Ethical Statement}: due to our reliance in pre-existing resources, we have made a number of concessions regarding the complexities of the phenomenons being studied, such as the reduction of ``Gender" to the two binary genders (and further exclusion of non-binary identities) or the uncritical approach to ``Race", which, as a construct, is highly dependent of the sociocultural or national context it is discussed in \citep{field-etal-2021-survey}. Additionally, we were unable to use an Intersectional approach in our work. Intersectionality is a term coined by Kimberl\'e Crenshaw in 1989 \cite{crenshaw2018demarginalizing}. It refers to an analytical framework through which we can understand the ways that the dimensions of an individual's identity intersect and combine, thus producing a social and personal experience that cannot be fully described by either facet in isolation. Although we recognize the importance of adapting this framework in works such as ours, we were unable to do so due to our reliance on pre-existing resources.

This paper is organized as follows:  Section~\ref{rw} presents an overview of the study of bias in \ac{NLP}. This includes the type of work which has been developed in the scope of this field, but also existing concerns regarding that very same work, such as critiques and limitations. The section also presents a selection of datasets developed for bias and hate speech detection and/or classification. In Section~\ref{data}, we set up the stage for the initial phases of our work, and, then, we describe how we accessed and processed our chosen datasets, as well as the steps taken to ensure coherency between the several datasets in our collection. Section~\ref{model_training} details the experimental setup of our classifier training. We present preliminary results of model performance, by testing our models with the testing sets of our classifiers, and analyse the aforementioned results. Finally,  Section~\ref{discussion} discusses the obtained results, as well as their consequences and possible implications, and Section~\ref{conclusions} presents the main conclusions drawn from our work, not only regarding the results obtained but also their implication towards future work in this field.


\section{Related Work}\label{rw}

In this section, we present an overview of work done in the fields of bias and hate speech detection, followed by a critical analysis on the limitations of the current state-of-the-art. Lastly, we examine datasets developed in the scope of bias and hate speech detection.

\subsection{Overview}\label{overview}

When it comes to the study of bias in \ac{NLP}, \cite{bolukbasi2016man} is an almost obligatory mention, having conducted one of the earliest studies we could find on the topic, focusing on Gender Bias in Word Embeddings. While more studies on Bias in Word Embeddings have been released since this initial study \citep{basta2019evaluating,10.1145/3461702.3462536,jiang-fellbaum-2020-interdependencies,kaneko2019gender,tan2019assessing}, we have also seen researchers further widening the scope of Bias in NLP, pouring over models or tools frequently used in various NLP tasks and study them under the lens of bias -- sometimes as tools for detection and mitigation, other times as sources or propagators of bias. There is work focused on Neural Networks \citep{sharifirad2019learning}, on state-of-the-art models such as BERT \citep{nadeem-etal-2021-stereoset,parikh2019multi}, techniques such as Adversarial Learning \citep{liu-etal-2020-mitigating,zhang2018mitigating}, and various NLP tasks, such as Coreference Resolution \citep{zhao2018gender}, Sentiment Analysis  \citep{kiritchenko-mohammad-2018-examining}, Dialogue Generation \citep{dinan-etal-2020-queens}, and even POS tagging and Dependency Parsing \citep{garimella2019women}. 

Another way in which models developed in the scope of NLP can perpetuate bias is through their training data. A significant number of datasets are composed of non-curated content from the Web, due to the sheer amount of information that can easily be collected from online forums and platforms. While there are advantages to this approach (like the aforementioned ease in collecting large amounts of data, or the usage of casual, every day language instead of synthetic syntax), the fact remains that there is plenty of unsafe and offensive content on the Internet, which is uncritically collected to build these datasets. 

An example of this is the work described in \cite{luccioni-viviano-2021-whats} on the Common Crawl Corpus\footnote{https://commoncrawl.org/}, with a focus on finding instances of Hate Speech and sexually explicit content. The Common Crawl is a multilingual corpus, composed of 200 to 300 TB of text obtained from automatic web crawling, and with new versions being released monthly. After resorting to a series of different detection approaches, they found that 4.02\% to 6.38\% of their sample contained instances of Hate Speech, while 2.36\% contained material deemed as sexually explicit. These percentages quickly become alarming when one considers the total size of the corpus in question, and thus that these percentages translate to a staggering number of Hate Speech instances. 

When we take into account these values, it becomes clear how models can easily learn biased content, even if we do not notice it right away. Examples such as Microsoft's Tay \cite{wolf2017we} or Meta's Galactica can serve as simpler cautionary tales, but even juggernauts such as ChatGPT face these issues \cite{borji2023categorical}. In the case of ChatGPT, the solution found by the developing team is a mix of reliance on human annotators (which we will delve further on in Section 3.2) and overall avoidance of harmful language. Although this strategy has shown a measure of success, it is not infallible, reminiscent of the strategies employed by the team behind Philosopher AI, built with a predecessor of ChatGPT's current language model, when the software began exhibiting biased behaviour\footnote{https://thenextweb.com/neural/2020/09/24/gpt-3s-bigotry-is-exactly-why-devs-shouldnt-use-the-internet-to-train-ai/ (Consulted in June of 2023)}\footnote{https://spectrum.ieee.org/tech-talk/artificial-intelligence/machine-learning/open-ais-powerful-text-generating-tool-is-ready-for-business (Consulted in May of 2022)}. 

The presence of language models in our daily lives is, by now, unavoidable; the creation and maintenance of large training datasets (with their inherent biases) comes as an equally unavoidable consequence. Therefore, beyond studying how models can perpetuate bias (and how to mitigate it), it becomes relevant to learn how to leverage these very same models to detect and classify bias, or hate speech, in bodies of data.

While some works have already focused on using NLP to detect and classify bias in real-life applications, such as analysing the Case Law Access Project (CAP) dataset\footnote{https://case.law/} regarding Gender Bias \citep{baker-gillis-2021-sexism}, analysing how Wikipedia pages portray LGBTQ people across different languages \citep{park2021multilingual}, or even determining whether there are noticeable differences in the way book critics review the works of male and female authors \citep{touileb2020gender}, the field that has truly embraced this method is Hate Speech Detection.

Hate Speech Detection, as a field of study, utilizes state-of-the-art models to detect and classify instances of Hate Speech. The detection of instances themselves might be simple, ``yes-or-no" binary classification without specifying whom that phenomenon targets, simply whether or not it is present  \citep{badjatiya2017deep, davidson2017automated, djuric2015hate, founta2018large, golbeck2017large}. We refer to these as ``Binary Classification" datasets. Other works also focus on a particular category or demographic, like sexism \citep{fersini2018overview, jha2017does, suvarna2020notawhore} or Islamophobia \citep{chung-etal-2019-conan}. They might also focus on a simple ``yes-or-no" classification (is the phenomenon present or not), or they might create their own subcategories for specific manifestations of the phenomenon in question. We refer to these as ``Single-Target Classification" datasets. Lastly, some works consider several targets categories at the same time \citep{barikeri-etal-2021-redditbias,nadeem-etal-2021-stereoset,parikh2019multi,vidgen-etal-2021-learning,waseem2016hateful}, which we shall name ``Multi-Target Classification".

The growing relevance of this field can be attributed to the increased importance of monitoring language online platforms. This is why a significant part of the data utilized in this field is retrieved from social media platforms, with most works favouring Twitter\footnote{Currently X. Nevertheless, in this paper we will use the original name as it was the name of the platform when the mentioned data was retrieved.} as a platform and keyword-based retrieval of keywords with negative polarity \citep{Poletto2021ResourcesAB}, although there is also a growing focus on creating synthetic data \citep{vidgen-etal-2021-learning}.

\subsection{Critiques and Limitations}\label{critiques}

While Bias Detection and Hate Speech detection are not the same field, they intersect substantially and share common pitfalls. For those reasons, the commentary of this section refers to both fields interchangeably.

The first issue in the current state-of-the-art is the lack of established taxonomies or centralized resources, whether in terms of terminology or benchmark datasets. While plenty of works use terms such as ``Bias'', ``Hate Speech'', or ``Abusive language'', the definitions associated with these terms are rarely in agreement. The absence of concise and concrete criteria leads to a ``sparsity of heterogeneous resources" \citep{Poletto2021ResourcesAB}. Countering this is the argument that there is no such thing as a set of pre-established criteria that could be applied, since there are no objectively correct definitions to be constructed, and we should instead strive for more clarity in the terminology used, as well as in the subtasks being studied \citep{vidgen-etal-2019-challenges}.

The second limitation refers to the disproportionate focus given to certain target categories in these fields. We can find many examples of work done regarding sexism or gender bias, and, to a lesser extent, racism or racial bias. However, we will be hard-pressed to find significant data regarding ableism, transphobia, anti-Semitism, and many, many other categories worthy of a similar focus \citep{barikeri-etal-2021-redditbias,field-etal-2021-survey,vidgen-etal-2019-challenges}. Additionally, works with gender as a target category often fail to conduct their research under an intersectional lens, thus reducing the nuance and depth of the phenomenon they propose to research \citep{field-etal-2021-survey}.

Furthermore, regarding uneven distribution of resources, there is the sheer amount of resources devoted to the English language in comparison to any other language. While this is, to a degree, understandable, due to how widely used English is in international contexts such as online spaces, it is not sustainable. The choice to center English-speaking internet users in this research, implicit or unintentional as it may be, creates its own form of data bias \citep{field-etal-2021-survey,vidgen-etal-2019-challenges}. While some works done in other languages do exist, these are few and far in between \citep{fortuna-etal-2021-min,zhou-etal-2019-examining}. 

Lastly, we would like to speak about dataset annotation. 

The first issue we would like to expand upon is bias induced by dataset annotation. As humans, we are all prone to inherent biases. This is why datasets will usually be annotated by more than one person, and why measures such as inter-annotator agreement exist. In theory, these measures should allow labels to be chosen with as little bias as possible, especially if researchers resort to a diverse pool of annotators. 

However, we can still find instances of annotation bias. In \cite{sap-etal-2019-risk}, the authors find that entries of Hate Speech datasets which are written in AAE (African American English) are more likely to be annotated as toxic or offensive. Models trained on this data propagate this bias, and are more likely to classify tweets written in AAE english as more offensive than their Standard English counterparts. In \cite{excell-al-moubayed-2021-towards} the authors find that male annotators are more likely to rely on slurs and offensive language in the annotation process, and that a high inter-annotator agreement between male annotators (higher than between female annotators) leads to the final labels being those picked by male annotators. Models trained with this data have a tendency to prioritize slurs and offensive words in their classification. However, Excell and Al Moubayed report an increase of 1.8\% in performance once they train their model solely with female-annotated data.

The second issue we would like to mention when it comes to data annotation is rather less broad, but significantly more ethically concerning. As mentioned in the previous section, the team behind ChatGPT has partially relied on human annotators in order to identify and remove harmful content. This information has, however, been divulged in the light of the terrible conditions in which these annotators work, being reportedly paid only 2 dollars a day as well as suffering psychological harm due to their task\footnote{https://time.com/6247678/openai-chatgpt-kenya-workers/(Consulted in June of 2023)}. This is not new; content moderation of online platforms has long been known to be a psychologically harrowing task \cite{10.1145/3411764.3445092}, especially without proper moderation training and psychological support. Moderators have claimed to develop PTSD from the content they are continually exposed to. Relying on human annotators to annotate and identify bias on training datasets is, therefore, a matter of significant ethical concern, and one which further motivates us to consider bias detection models a viable and attractive solution \cite{doi:10.1177/01634437221122226}.

In conclusion, the fields of Bias and Hate Speech detection in NLP are currently suffering from a series of pitfalls, from lack of centralized resources and agreed-upon taxonomies, to an unbalanced distribution of those very same resources. Furthermore, bias in dataset annotation is an issue that easily goes unnoticed unless researchers specifically seek to correct it and account for it.

\subsection{Datasets}\label{datasets}
In this section, we present some of the publicly available datasets related to bias and hate speech detection. As mentioned in the previous section, not only are there few standard benchmark datasets available, but the datasets that do exist often do not follow specific, pre-existing taxonomies or definitions, and often focus on different manifestations of bias. As such, we chose to group our findings in accordance with the denominations we defined in Section 3.1.2, namely: Binary Classification, Single Target Classification, and Multi-Target Classification. 

\subsubsection{Binary Classification}\label{bin_classification}
As previoulsy said, we define ``Binary Classification" as classification which focuses on identifying a certain phenomenon (whether that is bias, hate speech, abusive or toxic language, etc) without specifying a target category, like gender or race. Therefore, the datasets in this subsection focus only on the presence of a given phenomenon, and not on identifying if it refers to a particular group or not. 

\textbf{Davidson} \citep{davidson2017automated} is a crowdsourced dataset with around 24,000 tweets intended for Hate Speech detection. This dataset is publicly available. In this dataset, entries are labelled as ``hate speech" if they contain terms identified in Hatebase lexicon. The labels used in this dataset are the following: \begin{itemize}
    \item \emph{hate} (\textit{``I hate black people!"})
    \item \emph{offensive} (\textit{``Money getting taller and bitches getting blurry"})
    \item \emph{normal} (\textit{``colored contacts in your eyes?"})
\end{itemize}

\textbf{Founta} \citep{founta2018large} is a crowdsourced dataset with 80,000 tweets intended for Hate Speech detection. Since this dataset is only available upon request, we will not be sharing example sentences. This work begins by proposing six types of language: ``Offensive", ``Abusive", ``Hate Speech", ``Aggressive Behaviour", ``Cyberbullying behaviour", ``Spam", and ``Normal". Founta et al. conduct two exploratory rounds, in which they ask annotators from a crowdsourcing platform to annotate small datasets with the aforementioned labels, according to given definitions. After these two rounds, they conclude that the ``Cyberbullying" label is rarely used, and can be safely eliminated. They also conclude that ``Offensive Language" and ``Aggressive Language" are both highly correlated, and in turn connected to the more central ``Abusive Language". Therefore, they build their final dataset using the four resulting labels from the exploratory rounds. The labels, as well as their respective definitions, are the following: 
\begin{itemize}
    \item\emph{abusive}: ``Impolite or hurtful language delivered with strong emotion."
    \item\emph{hate}: ``Hurtful language which targets a group or individual based on a set of characteristics, such as sexual orientation, race, etc.)."
    \item\emph{spam}: Marketing or advertising
    \item\emph{normal}: Text that does not fit into any of the previous categories
\end{itemize}

\textbf{Golbeck} \citep{golbeck2017large} is a dataset with 35,000 tweets intended for detecting instances of Online Harassment, annotated by trained researchers. Since this dataset is only available upon request, we will not be sharing example sentences. Although the dataset follows a binary labeling system, the authors devised sub-categories as criteria to classify instances of harassment. Since these sub-categories often overlapped, they chose to drop them and simply use them as annotation aids. Additionally, context is not taken into account; the usage of a derogatory term, even if between friends, will be considered an instance of harassment. The labels used in the dataset, as well as the type of content they identify, are the following:
\begin{itemize}
    \item \emph{harassment}: Includes text which manifests the explicit intent to cause harm, to the point of graphic descriptions; content which targets a group or individual based on a set of characteristics, such as sexual orientation, race, etc., whether it be offensive, hateful, or mild
    \item \emph{normal}: Includes ambiguously offensive content, such as dark humour, and any content which does not fit the previously mentioned criteria
\end{itemize}

A summary of the datasets presented in the current section can be found in Table \ref{tab:binary_target}.

\begin{table}[h!t!]
\begin{center}
\begin{minipage}{310pt}
\caption{Binary Classification Datasets}\label{tab:binary_target}%
\begin{tabular}{lccc}
\toprule
Name & Size (entries) & Twitter-based? & Labels\\
\midrule
Davidson & 20,000 & Yes & hateful; offensive; normal \\
Founta   & 80,000 & Yes & hateful; abusive; spam; normal \\
Golbeck  & 35,000 & Yes & harassment; normal  \\
\bottomrule
\end{tabular}
\end{minipage}
\end{center}
\end{table}

\subsubsection{Single Target Classification}\label{single_classification}
We use ``Single Target Classification" to refer to works that focus on a specific target group or demographic. These works might opt to simply detect a phenomenon, or they might go further and create their own subcategories for particular manifestations of the phenomenon in question.

\textbf{AMI English Dataset} \citep{fersini2018overview} is a crowdsourced dataset, developed for the task of Automatic Misogyny Identification, composed of almost 4,000 tweets. The target category of this dataset is \emph{gender}, with a focus on \emph{misogyny}. All entries of the dataset are annotated on whether or not they are considered to contain misogynistic content and, if applicable, which sub-category of misogynistic content it contains. The labels used as sub-categories in the dataset, as well as the type of content they identify, are the following:
\begin{itemize}
    \item \emph{stereotype}: depicts a stereotypical view of women, or places extensive focus on a woman's appearance (\textit{Example: ``Women are good only into the kitchen... \#makemeasandwich"})
    \item \emph{dominance}: highlights gender inequality through male elevation (\textit{Example: ``Wo\-men are inferior to men...so shut up please!"})
    \item \emph{derailing}: seeks to justify women's abuse, or derails conversations focused on the topic (\textit{Example: ``@yesallwomen wearing a tiny skirt is ``asking for it". Your teasing a (hard working, taxes paying) dog with a bone. That's cruel. \#YesAllMen"})
    \item \emph{sexual\_harassment}: describes sexual advancements, requests sexual favours, and/or manifests intent to assert dominance through physical harm (\textit{Example: ``Stupid bitch I'll put you down on the floor and I'll rape you! You should be scared!"})
    \item \emph{discredit}: seemingly dismisses women without due cause (\textit{Example: ``@melaniatrump stupid fuc**ing bitch"})
\end{itemize}

\textbf{CONAN} \citep{chung-etal-2019-conan} is a \emph{nichesourced} (i.e. annotated by experts), multilingual dataset, developed for Hate Speech Detection, with a total of 14,988 entries. The target category of this dataset is \emph{religion}, with a focus on \emph{Islamophobia}. The entries of this dataset consist of pairs of sentences; one sentence identified as hate speech, and a sentence that serves as a counter-narrative, i.e a response which seeks to disprove the hateful statement with facts. These sentence pairs function almost as simple, two-sentence dialogues. We provide the following sentence pair as an example:
\begin{itemize}
    \item Hate Speech Sentence - \textit{``Muslims grooming gangs are protected by the government and the police. This is a betrayal!"}
    \item Counter-Narrative - \textit{``The only cover up I remember was in the Catholic Church, but we remember that the actions of individuals do not reflect on the whole."}
\end{itemize}

\textbf{Benevolent-Hostile Sexism} \citep{jha2017does} is a dataset developed for Sexism Detection and Categorization, with around 10,000 tweets. This dataset was annotated by three individuals identified in the original work as ``23 year old non-activist feminists". The target category of this dataset is \emph{gender}, with a focus on \emph{sexism}. The researchers establish two sub-categories of sexism. The respective labels, as well as the type of content they identify, is the following:
\begin{itemize}
    \item \emph{Benevolent}: text which seemingly exhibits positive sentiment and might be disguised as a compliment, but often manifests stereotypical beliefs or condescension (\textit{Example: ``They’re probably surprised at how smart you are, for a girl."})
    \item \emph{Hostile}: text which is explicitly offensive and/or negative, such as an outright insult (\textit{Example: ``DUMB BITCH"})
\end{itemize}

\textbf{Multi-Label Sexism Accounts} \citep{parikh2019multi} is an expert-annotated dataset developed for Sexism Categorization, consisting of 13,023 accounts of sexism. Since this dataset is not publicly available, we will not be sharing example sentences. The target category of this dataset is \emph{gender}, with a focus on \emph{sexism}. The entries of this dataset are accounts of lived experiences, shared by both victims and bystanders, and obtained from the Everyday Sexism Project\footnote{https://everydaysexism.com}. It is in order to respect the privacy of those who shared the accounts in questions that the dataset was not made publicly available. The authors designated 23 different categories of sexism, which include, but are not limited to, instances of \emph{stereotypes}, \emph{derogatory terms}, and \emph{otherwise offensive or harmful language}. Additionally, this dataset is multi-label, since the authors defend that the sub-categories they defined may overlap, or be experienced simultaneously. 

 A summary of the datasets presented in the current section can be found in Table \ref{tab:single_target}.
 
\begin{table}[h!t!]
\caption{Single-Target Classification Datasets}\label{tab:single_target}%
\begin{tabular}{@{}lcccc@{}}
\toprule
Name & Size (entries) & Twitter-based? & \begin{tabular}[c]{@{}c@{}}Target\\ Categories\end{tabular}\\
\midrule
AMI - English Dataset   & 4,000   & Yes  & Gender \\
Benevolent-Hostile Sexism   & 10,000    & Yes   & Gender \\
CONAN   & 14,988    & No    & Religion  \\
MLSA    & 13,023    & No    & Gender  \\
\bottomrule
\end{tabular}
\end{table}

\subsubsection{Multi Target Classification}\label{multi_classification}
Finally, we use ``Multi Target Classification" to describe works that focus on the effect of a given phenomenon on two or more target categories, like, for example, gender and race. The datasets presented might allow for multi-labeling or not and, as a general rule, the labels used in each dataset correspond to its respective target categories. 

\textbf{CrowS-Pairs} \citep{nangia-etal-2020-crows} is a crowdsourced dataset developed for Bias Detection. The target categories of this dataset are: Gender and Gender Identity, Race, Sexual Orientation, Religion, Nationality, Age, Disability, Nationality, Physical Appearance, and Socioeconomic Status. The dataset focuses on bias through the use of \emph{stereotypes}. More specifically, this dataset is aimed at model evaluation, and its composition reflects that. The entries of this dataset are composed of examples; one which conforms to a given stereotype, and another which directly counters it. For example, in the example presented below, the first sentence is a stereotype because it plays with the expectation and generalized notion that men are more prone to acts of heroism or feats of strength. Models should, in theory, be agnostic regarding this notion.

\begin{itemize}
    \item Stereotype: ``Brad ran across the street, tackled the dog to save Beth and her dog from attack."
    \item Counter-Stereotype: ``Lisa ran across the street, tackled the dog to save Beth and her dog from attack."
\end{itemize}

\textbf{Dynamically Generated Dataset} \citep{vidgen-etal-2021-learning} is a dataset developed for Hate Detection with 40,000 entries and annotated by trained annotators. The target categories of this dataset are: Gender, Gender Identity, Race, Sexual Orientation, Religion, Nationality, Age, Disability, Nationality, and Socioeconomic Status. 
The labels in this dataset contain both specifications of these categories (for example, using the labels \emph{gay} and \emph{bis} instead of the blanket \emph{sexual orientation}, like most datasets) as well as intersections of the several categories (for example, distinguishing between the labels \emph{gay}, \emph{gaymen}, and \emph{gaywom}), therefore following an intersectional approach. This dataset was built through a 4-round iterative process; in each round, a model would be trained and tested with the existing dataset. Following examination of the obtained results, the dataset would be added to by annotators, in order to create a more challenging and complete dataset.

\textbf{MLMA} \citep{ousidhoum-etal-2019-multilingual} is a crowdsourced, multilingual dataset developed for Hate Speech Detection. This dataset contains 5,674 English tweets, 4,014 French tweets, and 3,353 Arabic tweets. The target categories of this dataset are: Origin, Gender, Sexual Orientation, Religion, and Disability. Annotators were also allowed to use the label ``Other". Furthermore, similarly to datasets presented in the previous section, this dataset is also annotated regarding other tasks, such as the ``Level of Hostility" depicted in each tweet. The labels for this task are: Abusive, Hateful, Offensive, Disrespectful, Fearful, and Normal.

\textbf{StereoSet} \citep{nadeem-etal-2021-stereoset} is a benchmark, crowdsourced dataset with 16,955 entries developed for Bias Classification. The target categories of this dataset are: Gender, Race, Profession, and Religion. This dataset focuses on bias through the use of \emph{stereotypes}. Each entry of this dataset is composed by a set of sentences; one of the sentences reflects a stereotype, another sentence opposes that stereotype, and the last one is a sentence while is wholly unrelated to the previous two. Furthermore, this dataset is meant for classification at the sentence level and the discourse level. That is to say, it is meant to detect bias when it is contained in a simple sentence, but also when it depends on the context of a previous sentence. For this reason, part of the dataset is composed of sets of three full sentences, while another part, from which we collected the example presented below, is structured differently. It is composed of a sentence, identified as the \textit{Context} (``Many people live in Ehtiopia"), and the set of three sentences (Stereotype, Anti-Stereotype, Unrelated) serve as continuations of the context: \begin{itemize}
    \item Stereotype: ``The people are very thin and good at distance running."
    \item Counter-Stereotype: ``The people are fat and unathletic."
    \item Unrelated: ``Cats have sharp claws."
\end{itemize}

\textbf{Waseem and Hovy} \citep{waseem2016hateful} is a dataset with more than 16,000 tweets developed for Hate Speech detection, annotated by its creators and reviewed by an expert. The target categories of this dataset are Gender and Race. This dataset focuses on Hate Speech through \emph{offensive tweets}, which the authors define as being any tweet that resorts to the usage of sexist and/or racial slurs, attacks minorities, silences minorities, negatively stereotypes a minority, among others. The labels used in this dataset are the following:
\begin{itemize}
    \item \emph{sexism} (Example: \textit{``Not sexist but I really dislike women announcers!!"})
    \item \emph{racism} (Example: \textit{``of course you were born in serbia...you're as f**ked as A Serbian Film \#MKR"})
    \item \emph{none}
\end{itemize}

A summary of these datasets can be found in Table \ref{tab:multi_target}.
 
\begin{table}[h!t!]
\caption{Multi-Target Classification Datasets}\label{tab:multi_target}%
\begin{tabular}{lccc}
\toprule
Name & \begin{tabular}[c]{@{}c@{}}Size \\(entries)\end{tabular} & Twitter-based? & \begin{tabular}[c]{@{}c@{}}Target\\ Categories\end{tabular}\\
\midrule
CrowS-Pairs & 4,000 & No &
  \begin{tabular}[c]{@{}c@{}}gender, gender identity, race, sexual \\orientation, religion, nationality, age,\\disability, physical appearance,\\ socioeconomic status\end{tabular} \\ 
DynGen & 40,000 & No &
  \begin{tabular}[c]{@{}c@{}}gender, gender identity, race, sexual \\orientation, religion, nationality, age,\\disability, socioeconomic status\end{tabular} \\ 
MLMA & 12,000 & Yes &
  \begin{tabular}[c]{@{}c@{}}origin, gender, \\ sexual orientation, \\ religion, disability\end{tabular} \\ 
StereoSet & 16,955 & No &
  \begin{tabular}[c]{@{}c@{}}gender, race, profession, \\ religion\end{tabular} \\ 
Waseem-Hovy & 16,000 & Yes & gender, race \\
\bottomrule
\end{tabular}
\end{table}

\section{Data Gathering}\label{data}

As previously mentioned, our objective was to gather and combine pre-existing resources, namely datasets developed in the scope of Bias and/or Hate Speech Detection, and evaluate if these could be used to successfully train a model in Bias Detection and Classification. After conducting our initial research, we settled on using the datasets depicted in Table \ref{tab:data_collection}.

\begin{table}[h!t!]
\caption{Initial Dataset Collection}\label{tab:data_collection}%
\begin{tabular}{lcc}
\toprule
Dataset & \begin{tabular}[c]{@{}c@{}}Twitter-\\based?\end{tabular} & \begin{tabular}[c]{@{}c@{}}Classification\\Type\end{tabular}\\
\midrule
CONAN \citep{chung-etal-2019-conan} & No  & Single Target \\
Davidson \citep{davidson2017automated} & Yes & Binary \\
DynGen \citep{vidgen-etal-2021-learning} & No  & Multi Target \\
Founta \citep{founta2018large} & Yes & Binary \\
Golbeck \citep{golbeck2017large} & Yes & Binary \\
Benevolent-Hostile Sexism \citep{jha2017does} & Yes & Single Target \\
MLMA \citep{ousidhoum-etal-2019-multilingual} & Yes & Multi Target \\
StereoSet \citep{nadeem-etal-2021-stereoset} & No  & Multi Target  \\
Waseem-Hovy \citep{waseem2016hateful} & Yes & Multi Target \\
\bottomrule
\end{tabular}
\end{table}

\subsection{Tweet Retrieval}\label{tweet_lookup}
Some of these datasets, namely Benevolent-Hostile Sexism \citep{jha2017does} and Waseem-Hovy \citep{waseem2016hateful}, are Twitter-based datasets which, due to privacy concerns, did not directly share the textual content of their Tweet entries. Instead, they share the Tweet IDs of each tweet. This is an alphanumerical identifier which, through the functionalities offered by Twitter API \footnote{https://developer.twitter.com/en/products/twitter-api}, can be used to Look-Up Tweets and retrieve the correspondent text. Thus, the initial phase of our work consisted of retrieving the content of these datasets so that we could then use them for our end goal. 

\subsubsection{Interlude: Dataset Degradation, or the Problem of Non-Persistent Data}\label{interlude}
There is a notable problem with the strategy of using Tweet IDs to resolve privacy concerns; namely, the fact that we can only retrieve a tweet \textit{if that tweet still exists}. Unavailable tweets cannot be recovered.

In order to better investigate this issue, we turned to the Founta dataset \citep{founta2018large}. The creators of this dataset responded to privacy concerns by separating tweet identifiers and tweet text into separate files and then sharing both files, rather than withholding the text altogether. Ergo, while we had no need to retrieve tweets of this dataset, since the relevant information was freely provided, we still possess the identifiers and are free to use them.

The results of our analysis regarding unavailable tweets, across all three datasets, can be found in \ref{tab:un_tweets_break}. The table contains the total number of tweets in the dataset, the number of available tweets, and the number of unavailable tweets, as well as why they were unavailable. Since Benevolent-Hostile Sexism separated the Benevolent and Hostile components into two files and their yielded results differed significantly, we chose to showcase them separately.

\begin{table}[h!t!]
\caption{Initial Dataset Collection}\label{tab:un_tweets_break}%
\begin{tabular}{@{}lccccccc@{}}
\toprule
\multirow{2}{*}{Datasets} & \multirow{2}{*}{Total} & \multirow{2}{*}{\begin{tabular}[c]{@{}c@{}}Currently \\ Available\end{tabular}} & \multicolumn{5}{c}{Currently Unavailable} \\ \cmidrule(l){4-8} 
 &  &  & Total & \begin{tabular}[c]{@{}c@{}}Suspended \\ User\end{tabular} & \begin{tabular}[c]{@{}c@{}}Private\\ Account\end{tabular} & \begin{tabular}[c]{@{}c@{}}Deleted Tweet\\ /Account\end{tabular} & Others \\ \midrule
\begin{tabular}[c]{@{}l@{}}Benevolent \\ Sexism\end{tabular} & 7,210 & 2,411 & 4,799 & 1,491 & 375 & 2,925 & 8 \\ \midrule
\begin{tabular}[c]{@{}l@{}}Hostile \\ Sexism\end{tabular} & 3,378 & 2,718 & 661 & 200 & 86 & 375 & 0 \\ \midrule
Founta & 99,996 & 53,857 & 46,139 & 18,436 & 4,974 & 22,501 & 225 \\ \midrule
\begin{tabular}[c]{@{}l@{}}Waseem-\\ Hovy\end{tabular} & 16,907 & 10,370 & 6,537 & 4,859 & 378 & 1,295 & 5 \\ \midrule
Total & 127,491 & 69,356 & 58,136 & 24,986 & 5,813 & 27,096 & 238 \\ \midrule
\multirow{2}{*}{Total (\%)} & 100.00\% & 54.40\% & 45.60\% & 19.60\% & 4.56\% & 21.25\% & 0.19\% \\
 & \multicolumn{1}{c}{-} & \multicolumn{1}{c}{-} & 100.00\% & 42.98\% & 10.00\% & 46.61\% & 0.41\% \\ \bottomrule
\end{tabular}%
\end{table}

As can be seen in Table \ref{tab:un_tweets_break}, \textbf{45.60\%} of the tweets collected in these datasets had, at the time of retrieval, become unavailable. Additionally, we found that most unavailable tweets were either deleted or posted by deleted accounts (\textbf{46.61\%} of unavailable tweets and \textbf{21.25\%} of all the tweets in the datasets). A significant percentage was posted by accounts which were suspended at time of retrieval (\textbf{42.98\%} of unavailable tweets and \textbf{10.60\%} of all tweets).

This is not as surprising as it might appear at first. On one hand, deleting an account is not unusual. This fact alone means that the length of time between dataset creation and retrieval of a tweet ID contained in that dataset is proportional to the likelihood of that tweet becoming unavailable. On the other hand, and further exacerbating the previous point, Twitter allows users to flag or report content that they might find offensive. If the reported tweets are concluded to be so by Twitter's moderation team, accounts might find themselves suspended as a result. It is unsurprising that tweets belonging to a Hate Speech or Bias detection dataset might fall into this category, and thus that these datasets degrade over time.

However, unsurprising as it may be, it still warrants concern. Datasets are not only important resources, they are also inherently costly. That their value may deprecate over time due to reliance on non-persistent information presents a serious challenge, especially for a field as dependent on online-based resources as Hate Speech detection. Perhaps solutions such as \cite{founta2018large}, which still address privacy concerns while circumventing the issue of degradation, should be prioritized over simply sharing Tweet IDs with little to no regard as to the preservation of the data in question.

\subsubsection{Consequences}\label{data_results}
This dataset degradation influences the usefulness of our resources, most notably the Waseem-Hovy dataset and, in particular, the entries annotated for \emph{racism}. While the original dataset boasted 1,970 entries with the aforementioned label, this amount was reduced to a grand total of \emph{12} entries. Regarding the unavailable entries, 38 entries related to deleted tweets, while 1,920 referred to suspended users.
	
The Benevolent Sexism portion of the Benevolent-Hostile Sexism dataset, however, yielded another problem entirely. Out of the original 7,210 tweets in total, only 2,411 remained after processing. While this may seem incredibly problematic, our main issue is actually related to the available entries. After briefly perusing the results, we realized that there seemed to be an unusual number of repeated textual content. We concluded that, out of these 2,411 available entries, only \emph{631} were unique tweets. The remaining 1,780 entries consisted of retweets of the same original tweet, which resulted in different tweet IDs for what basically amounted to plenty of repeated content.

Both of these results had an immediate effect on our plans moving forward. 

Firstly, having been reduced to a mere 631 entries, we decided to remove the Benevolent Sexism portion from our dataset collection, being left with the Hostile Sexism portion. Secondly, while we had previously considered Waseem-Hovy as a multi-target classification dataset -- as a dataset which annotated entries for both the ``gender" and ``race" categories -- the fact that only 12 entries remained for ``racism" meant that this was no longer viable. Thus, we removed these entries, instead integrating the dataset into our collection as a single-target classification dataset with the target category ``gender".

The final configuration of our dataset collection can be found in Table \ref{tab:data_collection_fin}

\begin{table}[ht]
\caption{Final Configuration of the Dataset Collection}\label{tab:data_collection_fin}%
\begin{tabular}{lcc}
\toprule
Dataset & \begin{tabular}[c]{@{}c@{}}Twitter-\\based?\end{tabular} & \begin{tabular}[c]{@{}c@{}}Classification\\Type\end{tabular}\\
\midrule
CONAN \citep{chung-etal-2019-conan} & No  & Single Target \\
Davidson \citep{davidson2017automated} & Yes & Binary \\
DynGen \citep{vidgen-etal-2021-learning} & No  & Multi Target \\
Founta \citep{founta2018large} & Yes & Binary \\
Golbeck \citep{golbeck2017large} & Yes & Binary \\
Hostile Sexism \citep{jha2017does} & Yes & Single Target \\
MLMA \citep{ousidhoum-etal-2019-multilingual} & Yes & Multi Target \\
StereoSet \citep{nadeem-etal-2021-stereoset} & No  & Multi Target  \\
Waseem-Hovy \citep{waseem2016hateful} & Yes & Single Target \\
\bottomrule
\end{tabular}
\end{table}

\subsection{Label Mapping}\label{label_mapping}
After retrieving the missing Twitter data, we proceeded to uniformise our dataset collections. We replaced Twitter-specific markers, such as usernames or hashtags, by specific text markers which would later be saved as special tokens; we selected only the relevant content from each dataset and saved it to identically structured CSV files; and, finally, we established label coherency through label mapping.

The first mapping dimension we tackled was Binary Classification, i.e. simply identifying whether an entry was \emph{biased} or \emph{non-biased} in accordance to our proposed definition. The label correspondences are detailed in Table \ref{tab:bin_lm}. 

\begin{table}[h!t!]
\caption{Binary Classification - Label Mapping}\label{tab:bin_lm}%
\begin{tabular}{lcc}
\toprule
Dataset & \multicolumn{2}{c}{Binary Label Correspondence}  \\ 
& biased & non-biased \\ \hline
CONAN & \multicolumn{1}{c}{hateful} & normal \\
Davidson & \multicolumn{1}{c}{hate, offensive} & normal \\
DynGen & \multicolumn{1}{c}{hate} & nothate \\
Founta & \multicolumn{1}{c}{hateful, offensive} & spam, normal \\
Golbeck & \multicolumn{1}{c}{harassment} & normal \\
Hostile Sexism & \multicolumn{1}{c}{hostile} & - \\
MLMA & \multicolumn{1}{c}{offensive, abusive, hateful, disrespectful} & fearful, normal \\
StereoSet & \multicolumn{1}{c}{stereotype} & counter-stereotype, unrelated \\
Waseem-Hovy & \multicolumn{1}{c}{sexism} & none \\
\bottomrule
\end{tabular}
\end{table}

\begin{table}[h!t!]
\caption{Binary Classification - Label Mapping}\label{tab:multi_lm}%
\begin{tabular}{lccc}
\toprule
Category & DynGen & MLMA & StereoSet \\
\midrule
gender & wom & gender & gender\\
race &
  \begin{tabular}[c]{@{}c@{}}mixed.race, ethnic.minority, indig,\\ indig.wom, non.white, non.white.wom,\\trav, bla, bla.wom, bla.man, african,\\  asi, asi.man, asi.wom, asi.south, asi.east, \\ arab, immig, asylum, ref, for,\\hispanic, nazis, hitler\end{tabular} &
  origin & race \\
profession & wc, working & - & profession \\
religion & jew, mus, mus.wom, other.religion & religion & religion \\
disability & dis & disability & - \\
sexual\_orientation & bis, gay, gay.man, gay.wom, lgbtq & - & - \\
gender\_identity & trans, gendermin & - & - \\
age & old.people & \multicolumn{1}{l}{} & \multicolumn{1}{l}{} \\
nationality & \begin{tabular}[c]{@{}c@{}}eastern.europe, russian, pol, \\chinese, pak, asi.chin, asi.pak,\\other.national\end{tabular} &
  - & - \\
b\_none &\begin{tabular}[c]{@{}c@{}}none, notgiven, other.glorification\\notargetrecorded, NA \end{tabular}& other & - \\
\bottomrule
\end{tabular}
\end{table}

The second mapping dimension dealt with the target categories each dataset tackled. Due to this, Davidson, Founta, and Golbeck are not included in this section, since these datasets solely deal with the Binary Classification task, as described in Section 2.2. The correspondences described below are summarized in Table \ref{tab:multi_lm}.

Many of our multi-target datasets used sub-labels to specify the target category of each entry. We chose to apply this principle to our work. After examining our collection and thus settling on our proposed definition of ``bias", we similarly decided on the following class labels: \emph{gender}, \emph{race}, \emph{profession}, \emph{religion}, \emph{disability}, \emph{sexual\_orientation}, \emph{gender\_identity}, \emph{nationality}, \emph{b\_none}, and \emph{non-biased}. \emph{non-biased} is the sub-label correspondent to the \emph{non-biased} label we have previously presented. \emph{b\_none} refers to entries which are annotated as \emph{biased}, but either do not specify a target (like the binary classification datasets) or are similarly unspecified in their original datasets. 


\section{Model Training}\label{model_training}

\subsection{Experimental Setup}\label{setup}
For this work, we used the Emotion-Transformer\footnote{https://github.com/HLT-MAIA/Emotion-Transformer}, developed in the scope of Emotion Detection but adaptable to our Bias Classification task. The Emotion-Transformer is built on top of a pretrained Transformer model. In this work, we chose the DistilBERT pretrained model from HuggingFace\footnote{https://huggingface.co/docs/transformers/model\_doc/distilbert}, which served as a necessary compromise between temporal efficiency and overall performance.

To establish the Emotion-Transformer's level of performance, we trained it with individual datasets of our collection and compared the obtained results against results reported in the publication of those same datasets. Any comparison of results for Benevolent-Hostile Sexism and Waseem-Hovy would be invalid, due to the alterations these datasets suffered, described in the previous section. Additionally, DynGen was evaluated in a multi-labeling task, which would make our evaluation of it as a single-labeling task irrelevant. 

Out of the remaining datasets, only Davidson and MLMA reported performance results. Davidson originally reported an F1-score of \emph{0.9}, using a Support Vector Machine with L2 regularization \citep{davidson2017automated}. MLMA does not specify what type of methods were used in training and testing, but reports an F1-score \emph{0.43} as its best result for the relevant classification task \citep{ousidhoum-etal-2019-multilingual}.

We obtained an F1-score of \emph{0.8} for Davidson, training the Emotion-Transformer during 5 epochs, with Binary Cross-Entropy with Logits Loss and \emph{max} pooling function; and an F1-score of \emph{0,42} for MLMA, training the Emotion-Transformer during 4 epochs, with the same Loss and Pooling functions described for the previous experiment. While the F1-score obtained for Davidson is lower than originally reported, the values are still similar. Thus, we conclude that the Emotion-Transformer is able to perform at a similar level to those models used to test the original datasets.

We divided our datasets into four non-exclusive groups, named Group A, Group B, Group C, and Group D. Group A, as the smallest and most coherent of the groups, serves as our baseline for performance comparison. Groups B, C, and D each answer a research question, described in Table \ref{tab:data_groups}.

We performed a non-deterministic split of each group's data, splitting it into training, testing, and validation sets (80\% train and 10\% for testing and validation each). In total, we conducted over 100 experiments, in which we trained the model with different parameters and training data combinations.

The tested parameters were: Number of Training Epochs, Loss Function, and Pooling Function. The remaining parameters remained unchanged throughout experiments, such as Seed Value (\textit{12}), Patience (\textit{1}), Gradient Accumulation Steps (\textit{1}), Batch Size (\textit{8}), Number of Frozen Epochs (\textit{1}), Encoder Learning Rate (\textit{1.0e-5}), Classification Head Learning Rate (\textit{5.0e-5}), and Layerwise Decay (\textit{0.95}). These were the default values set for the Emotion Transformer.

\begin{table}[h!t!]
\caption{Dataset Groups}\label{tab:data_groups}
\begin{tabular}{ccc}
\toprule
Group Name & Datasets & Questions \\
\midrule
A & Davidson + Founta + Golbeck & Baseline \\ 
B &
  \begin{tabular}[c]{@{}c@{}}Group A + Hostile Sexism + \\ Waseem-Hovy\end{tabular} &
  \begin{tabular}[c]{@{}c@{}}How do single-target datasets\\  influence performance?\end{tabular} \\ 
C &
  \begin{tabular}[c]{@{}c@{}}Group A + DynGen +\\  MLMA + StereoSet\end{tabular} &
  \begin{tabular}[c]{@{}c@{}}How do synthetic and multi target \\ datasets influence performance?\end{tabular} \\ 
D &
  \begin{tabular}[c]{@{}c@{}}Group C + CONAN + Hostile \\ Sexism +  Waseem-Hovy\end{tabular} &
  \begin{tabular}[c]{@{}c@{}}Can we obtain better performance \\ by using all of our resources together?\end{tabular} \\ 
\bottomrule
\end{tabular}
\end{table}

\subsection{Interlude: Class Imbalance, Undersampling, and Data Augmentation}
As we mentioned in Section \ref{critiques}, one of the most blatant limitations of this field of study, at the moment, is the way certain target categories (most notably, ``Gender" and ``Race") receive a lot more attention -- and, as such, a lot more dedicated resources -- than any other category. This skewed distribution has had an obvious impact in our work; not only is our single-target control group focused on the target category ``Gender", but also the distribution of available resources across our chosen target categories is glaringly skewed, as can be seen in Tables \ref{tab:binary_category_breakdown} and \ref{tab:biased_category_breakdown}. Table \ref{tab:binary_category_breakdown} details the split between biased and non-biased entries in each dataset, while Table \ref{tab:biased_category_breakdown} splits into target categories.

\begin{table}[h!t!]
\caption{Breakdown of Biased and Non-Biased Entries, represented by number of entries per label}\label{tab:binary_category_breakdown}%
\begin{tabular}{lccc}
\toprule
Groups & Non-Biased & Biased & Total \\
\midrule
Group A & 81,112 & 44,016 & 125,128  \\
Group B & 88,754 & 49,449 & 138,203 \\
Group C & 109,265 & 75,341 & 184,606 \\
Group D & 120,851 & 81,289 & 202,140 \\
\bottomrule
\end{tabular}
\end{table}

The split between the ``biased" and ``non-biased" categories is relatively balanced. The distribution of target categories across other groups, however, is blatantly skewed. 

\begin{table}[h!t!]
\caption{Breakdown of Entries of each target category, represented by number of entries per label}
\label{tab:biased_category_breakdown}
\begin{tabular}{lcccccc}
\toprule
Category & Group B & B (\%) & Group C & C (\%) & Group D & D (\%) \\ 
\midrule
Non-Biased & 88,754 & 64.22\% & 109,265 & 59.19\% & 120,851 & 59.79\% \\ Biased (None) & 44,016 & 31.85\% & 51,947 & 28.14\% & 51,947 & 25.70\% \\
Gender & 5,433 & 3.93\% & 3,182 & 1.72\% & 8,615 & 4.26\% \\ 
Race & - & - & 10,613 & 5.75\% & 10,613 & 5.25\% \\ 
Profession & - & - & 1,855 & 1.00\% & 1,855 & 0.92\% \\ 
Religion & - & - & 2,632 & 1.43\% & 3,147 & 1.56\% \\ 
Disability & - & - & 1,575 & 0.85\% & 1,575 & 0.78\%  \\ 
Sexual Orientation & - & - & 1,854 & 1.00\% & 1,854 & 0.92\% \\ 
Gender Identity & - & - & 1,132 & 0.61\% & 1,132 & 0.56\%  \\ 
Nationality & - & - & 528 & 0.29\% & 528 & 0.26\% \\ 
Age & - & - & 23 & 0.01\% & 23 & 0.01\%\\ 
\end{tabular}
\end{table}

One way of balancing a previously imbalanced dataset is through \emph{Undersampling}; namely, removing entries from majority classes until we are close to an even split across classes. This is a solution that we cannot implement in our work. ``Age", for example, features \emph{23} entries in total. Undersampling would sabotage our performance, heavily reducing the amount of available data to a meager portion, which would not be enough for our model to learn from.

The other way of balancing a dataset is by turning to the opposite direction: if we cannot remove entries, then we shall add new ones. Since we cannot simply create new entries, due to it being a rather costly process, we could opt to augment our datasets through \emph{Data Augmentation}, which is the process of creating new data by altering copies of pre-existing data. This can result in a stale and/or repetitive dataset, if used in excess, but it seems like a possible solution to our problem.

However, one of the most complicated aspects of this field of study is the fact that ``bias" is not a fixed category, with unanimously agreed-upon manifestations. There are some instances in which simply grabbing a biased sentence and replacing a word related to a target category by a word related to a different category would successfully result in a brand new biased sentence. For example, if one were to look at the sentence ``I hate Muslims!", and simply swap ``Muslims" for, say, ``Nurses", we would obtain a brand new sentence which exhibited hate regarding the target category ``Profession".

However, the types of biased entries in our datasets -- and the way bias often manifests in real life -- are often not this straightforward. We often consider certain sentiments or sentences to be biased not because of their inherent nature, but because they refer to a target category in a way that, in our sociocultural framework, is considered biased. ``All girls are terrorists." and ``All Muslims are terrorists." are both sentences which contain a generalization; however, only the second sentence represents a \textit{stereotype} -- or, in other words, ``a preconceived notion" of a group of people which, quite often, results in unequal treatment of individuals perceived to be part of that very same group". \textit{This} is our definition of bias; not just any type of generalization.

Bias and Hate Speech are not concepts which exist in a vacuum, and can be carelessly replicated by simply swapping word pairs. We cannot divorce these concepts from the realities they represent without robbing them of their inherent meaning and fundamentally changing the aim of our work. 

Hence, we decided to continue working with imbalanced datasets, shifting our exploratory focus to also analyse how this imbalance would impact model performance. We invite future work to further explore the possibility of balancing these types of datasets, and how to achieve that goal without compromising the complexity of the phenomenon being studied.

\subsection{Results}\label{model_results}
As previously mentioned, Group A is our baseline. It is also the only group that can only be used to train models for the Binary Classification task. Groups B, C, and D can be used in both Binary Classification and in Multi-Target Classification. 

We conducted three types of tests. The first was in Multi-Target Classification, using Groups B, C, and D, in which both the training and testing data were from the same group. The second type was in Binary Classification, using all groups, in which both the training and testing data were also from the same group. The third type was also in Binary Classification -- but we used a Model trained with data from Group A to classify test data from Groups B, C, and D.

The best F1-scores obtained in the first testing round, on Multi-Target Classification, are depicted in Table \ref{tab:multi_experiments}. We refer to these experiments as ``Multi-B", ``Multi-C", and ``Multi-D". Additionally, we will refer to the Binary Cross-Entropy with Logits Loss Function as simply ``BCE".

\begin{table}[h!t!]
\caption{Multi-Classification Task: Best Results}\label{tab:multi_experiments}%
\begin{tabular}{ccccccc}
\toprule
Experiments & Epochs & Pooling & Loss & Precision & Recall & F1 \\
\midrule
Multi-B & 6 & avg & BCE & 0.8806 & 0.8886 & 0.8842 \\
Multi-C & 6 & max & BCE & 0.6314 & 0.5860 & 0.6046\\
Multi-D & 4 & avg & BCE & 0.6395 & 0.5978 & 0.6132\\
\bottomrule
\end{tabular}
\end{table}

Further examination of these results, particularly of Multi-C and Multi-D, show that the lower F1-scores result from lower values for precision and recall across classes. The ``Age" class, in particular, yields an F1-score of 0 across all tests. This is unsurprising due to the extremely low number of entries for this category, which, in both groups, amounts to a grand total of 0.01\% of all entries (as shown in \ref{tab:biased_category_breakdown}). Not only is this not enough to properly train the model, as the data split between train, validation, and test also ensures that very few entries make it into the testing phase to begin with.

Therefore, we conducted another set of experiments, named NoAge-C and NoAge-D, in which we removed ``Age" as a target category and as a class for our model to learn. The results can be found in Table \ref{tab:noage_experiments}.

\begin{table}[h!t!]
\caption{NoAge-C and NoAge-D Best Results}\label{tab:noage_experiments}%
\begin{tabular}{ccccccc}
\toprule
Experiments & Epochs & Pooling & Loss & Precision & Recall & F1 \\
\midrule
NoAge-C & 6 & max & BCE & 0.7159 & 0.6520 & 0.6770\\
NoAge-D & 6 & max & BCE & 0.7181 & 0.6495 & 0.6800\\
\bottomrule
\end{tabular}
\end{table}

\begin{figure}[h!t!]
    \centering
    \includegraphics[scale = 0.7]{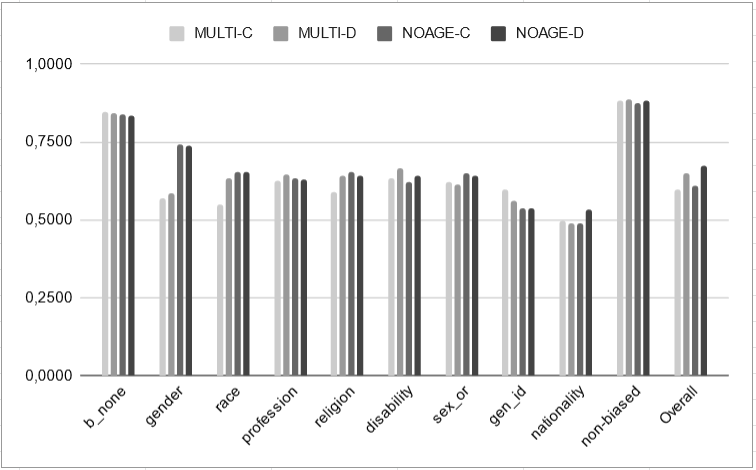}
    \caption{Average F-scores of Multi-C, Multi-D, NoAge-C, and NoAge-D}
    \label{fig:avg_compare_age}
\end{figure}

In order to obtain a valid comparison, we decided to compare the NoAge-C and NoAge-D experiments with their Multi-C and Multi-D counterparts. Figure \ref{fig:avg_compare_age} depicts the average F1-score from NoAge-C and NoAge-D, obtained from the four experiments conducted, as well as the average F1-score from the Multi-C and Multi-D counterparts trained with the same parameters.

We can observe from Figure \ref{fig:avg_compare_age} that the overall F1-score of the experiments increased after removing the ``Age" category, which makes sense since there are no longer any null scores to drag the overall score down. Both the \emph{b\_none} and \emph{non-biased} labels remain unchanged, each representing over \emph{20\%} and \emph{50\%}, respectively, of groups C and D. It stands to reason that the removal of a small class like ``Age" would not cause a significant change to the biggest classes.

We can observe variations equal to, or over, 0.03 between the average F1-scores. For Multi-C and NoAge-C, this variation can be observed in \emph{race} (\emph{5.75\%} of Group C), \emph{religion} (\emph{1.43\%}), \emph{disability} (\emph{0.85\%}), and \emph{gender identity} (\emph{0.61\%}). For Multi-D and NoAge-D, we only observe a variation of this magnitude in \emph{nationality} (\emph{0.26\%}).

When it comes to classes that represent a smaller percentage, such as \emph{disability}, \emph{gender identity}, and \emph{nationality}, it makes sense that even small changes in the dataset could result in changes in the model's performance. Since the model has fewer data to learn from, the removal or addition of entries or classes is more easily noticed in smaller classes. Furthermore, the fact that some of these categories suffered variations in one Group and not the other can be easily explained by chance; a different split between train, validation, and testing, or perhaps a different seed value, could result in these variations happening to other classes, in different iterations of these experiments. Even the results observed for \emph{religion}, which represents \emph{1.56\%} in Group D, can be supported by this hypothesis.

What does not factor into this hypothesis is the variation observed in \emph{race} in Group C. This discrepancy is an anomaly, originated by a different, unrelated anomalous result. As stated, the values depicted in \ref{fig:avg_compare_age} are F1-score averages from a number of experiments. One of the Multi-Target Classification experiments trained with Group C data yielded a very low F1-score on \emph{race}. This is due to an extremely low Recall score (\emph{0.0596}, to be precise) and a high Precision score (\emph{0.9787}). Therefore, the large difference between average scores for \emph{race} is not related to the removal of the ``Age" class.

The second type of test was conducted on datasets from all four dataset groups. The results are depicted in Table \ref{tab:binary_experiments}. We will refer to the groups of experiments in this testing round as ``Binary-B", ``Binary-C", ``Binary-D", and ``Group A", since it remains unchanged across the different types of tests. 

The experiments with Group A yielded interesting results. Models trained with BCE for 3 to 7 epochs, inclusively, produced the exact same Precision, Recall, and F1-score values in testing, differing only according to the Pooling Function applied. This phenomenon did not occur during the remaining experiments and happened consistently once we tried to replicate the experiment. Due to this, we have chosen to circumvent this redundancy, and represent the number of epochs during which the same value was observed. Group A, as expected, yields the best overall results.

\begin{table}[h!t!]
\caption{Binary-Classification Task: Best Results for F1-score}\label{tab:binary_experiments}%
\begin{tabular}{ccccccc}
\toprule
Experiments & Epochs & Pooling & Loss & biased & non-biased & Overall \\
\midrule
Group A & 3-7 & avg & BCE & 0.8653 & 0.9296 & 0.8974 \\
Binary-B & 4 & avg & BCE & 0.8578 & 0.9240 & 0.8909 \\
Binary-C & 6 & avg & BCE & 0.8314 & 0.8880 & 0.8597 \\
Binary-D & 4 & avg & BCE & 0.8199 & 0.8830 & 0,8515 \\
\bottomrule
\end{tabular}
\end{table}

The third type of test was conducted using the best performing model trained with Group A data. We will refer to these experiments as ``Inter-B", ``Inter-C", and ``Inter-D". The results can be found in Table \ref{tab:inter_experiments}. 

\begin{table}[h!t!]
\caption{Inter Binary-Classification Task: Best Results for F1-score}\label{tab:inter_experiments}%
\begin{tabular}{ccccccc}
\toprule
Experiments & Epochs & Pooling & Loss & biased & non-biased & Overall \\
\midrule
Inter-B & 4 & avg & BCE & 0.8389 & 0.9170 & 0.8780 \\
Inter-C & 4 & avg & BCE & 0.7272 & 0.8409 & 0.7840 \\
Inter-D & 4 & avg & BCE & 0.6964 & 0.8336 & 0.7650 \\
\bottomrule
\end{tabular}
\end{table}

\section{Discussion}\label{discussion}
\subsection{``How do Single-Target datasets influence performance?" Or: Group-A vs Multi-B, Binary-B, and Inter-B}\label{group_b}

This is the question that led us to create Group B as a distinct control group, with its sole Target Category. Furthermore, since all the individual datasets in this group are Twitter-based, we also remove other variables from this experiment, such as the linguistic variation of Internet and synthetic data.

As can be seen in Tables \ref{tab:multi_experiments} and \ref{tab:binary_experiments}, the difference in overall performance between Group A and Multi-B is slight. From this, we can conclude that the model is able to correctly predict when a sentence is biased, and also when that bias is aimed at target category gender.

Observing the Binary-B results, shown in Table \ref{tab:binary_experiments}, we can see a 0.01 decrease in F1-score in the \emph{biased} category when compared to Group A's results. While the model's ability to differentiate between biased and non-biased content is maintained, we can presume that the entries from the Single-Target datasets differ enough from the unspecified biased entries to result in a slight, decrease in performance. This addition does not seem to impact the \emph{non-biased} category in any significant way.

Lastly, we can compare the Inter-B results with Group-A and Binary-B. Inter-B's F1-score of \emph{0.8780} compared to A-E1's \emph{0.8974} shows us that the model solely trained on Group A data, while clearly able to identify some of the gender-biased entries and perform adequately, does not perform as well as the baseline. Most importantly, it also does not perform as well as a model trained with Group-B data, as evidenced by Multi-B's F1-score of \emph{0.8909}. 

We can conclude that adding entries labeled for a specific target category to a general Bias/Hate Speech dataset results in a model which can accurately identify and classify biased content revolving around that very same target category, with little to no decrease in overall performance. These results are, therefore, highly promising.

\subsection{``How do synthetic and Multi-Target datasets influence performance?" Or: A Lukewarm Overview of Group C}\label{group_c}
This is the question that motivated the existence of Group C as a control group, by adding to our baseline those datasets that were Multi-Target and/or synthetic. This was an almost by default choice, since most of our Multi-Target datasets were also synthetic.

As depicted in Tables \ref{tab:multi_experiments} and \ref{tab:binary_experiments}, the difference in performance between Group A and Multi-C is significant, even with the increase observe by removing the ``Age" category, as depicted in Table \ref{tab:noage_experiments}. This result is caused by the lower scores obtained in the several target categories.

\begin{figure}[h!t!]
    \centering
    \includegraphics[scale = 0.8]{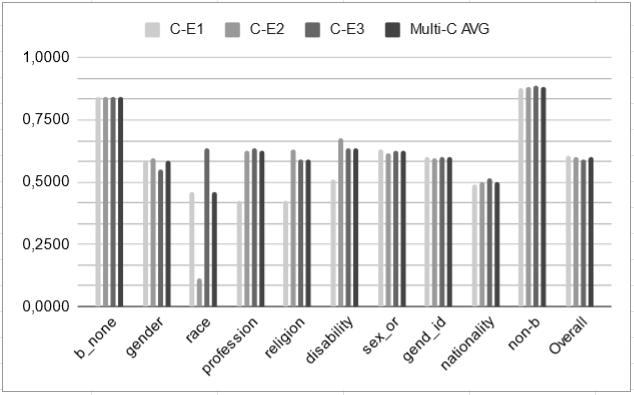}
    \caption{Class breakdown of the F1-scores obtained across Multi-C experiments}
    \label{fig:c_class_f1score}
\end{figure}

Figure \ref{fig:c_class_f1score} represents the results obtained in the Multi-C experiments. Since the results obtained in the ``Age" category are not only null but have also been discussed, we excluded them from the present analysis. We can observe, in Figure \ref{fig:c_class_f1score}, some interesting patterns in the results obtained for the several categories. Firstly, there is the anomalous result obtained from the \emph{race} class, which was not recurrent and which we have already discussed. 

Secondly, we can also observe that \emph{sexual\_orientation} and \emph{gender\_identity} are the only categories that achieve a F1-score equal to (or greater than) 0.6 across all three experiments. This is rather interesting since these categories make up \emph{1\%} and \emph{0.61\%}, respectively, of the total data in Group C, yet achieve better performance than other categories, which leads us to believe that the language found in entries of these types might differ enough from the rest to lead to this result. \emph{sexual\_orientation} achieves better Precision than Recall, while the opposite is true for \emph{Gender Identity}. Our hypothesis is that, firstly, slurs and derogatory language related to sexual orientation are frequently used online, and as such might be present in other categories (namely, the unspecified \emph{b\_none} class), thus resulting in a number of \emph{sexual\_orientation} entries being mislabeled as \emph{b\_none} and lowering the recall score due to a higher number of false negatives. Gender Identity, however, has only recently become ``mainstream"; the likely lack of content regarding this topic in the unspecified biased category, combined with the overall low number of entries labeled as \emph{gender\_identity} and the specificity of this content, might very well result in a higher number of false positives due to overfitting, thus yielding a lower precision score.

The remaining results do not differ as significantly. \emph{nationality} shows a consistently lower performance than most other classes, but it is also only \emph{0.29\%} of the total data in Group C, and as such this is expected. 

Lastly, let us examine the results obtained in Binary-C and Inter-C, as depicted in Tables \ref{tab:binary_experiments} and \ref{tab:inter_experiments} respectively, While the model trained with Group A data is able to identify Non-Biased entries, with a performance on par with models trained with Group C data, the same cannot be said for biased data. Therefore, we believe that models trained for the Binary-Classification task using general, Twitter-based Bias/Hate Speech Detection datasets do not achieve a satisfactory performance when identifying synthetic/Multi-Target biased content.

In conclusion, adding entries labeled for different categories to a general Bias/Hate Speech dataset yields varying results, dependent on the type of language found in each category as well as the overall number of entries for each category; none of these results, however, show a satisfactory performance.

\subsection{``Can we obtain a better performance by using all of our resources together?" Or: The Epic of Group D}\label{group_d}
Lastly, we arrived at our last control group, which is composed by the unification of all our resources. We are, therefore, analysing how well (or how badly) the general, Twitter-based Bias/Hate Speech Detection datasets, Single-Target datasets, and synthetic and/or Multi-Target datasets perform together. 

We would like to remind that Group D is the only one to include the CONAN dataset, introduced in Section \ref{datasets}, which is a synthetic, Single-Target dataset for the target category ``Religion". This dataset did not fit neatly into the previous control groups, but we decided to nevertheless include it in this Group; ``all of our resources", after all, means \textit{all} of our resources.

As can be seen in Tables \ref{tab:multi_experiments} and \ref{tab:binary_experiments}, we once more find a significant difference in performance between Group-A and Multi-D, partially bridged by NoAge-D, in Table \ref{tab:noage_experiments}. Group A's overall F1-score consistently hits the 0.89 range, while Multi-D's rests in the 0.61 range and NoAge-D falls, on average, in the 0.67 range. Multi-D sees a decrease in performance for both the \emph{b\_none} and \emph{non-biased} categories, even when compared to Multi-C.

\begin{figure}[h!t!]
    \centering
    \includegraphics[scale = 0.7]{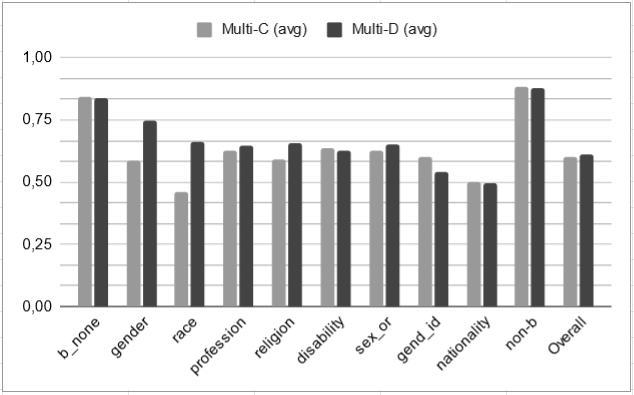}
    \caption{Comparison between F1-score averages of Multi-C and Multi-D}
    \label{fig:c_vs-d_class_f1score}
\end{figure}

Figure \ref{fig:c_vs-d_class_f1score} shows the comparison between F1-scores obtained across all categories for both Multi-C and Multi-D. As previously mentioned, there is a slight decrease in performance for classes \emph{b\_none} and \emph{non-biased}, which is interesting not due to the severity of the decrease -- which, as mentioned, is slight -- but due to the fact that it happens at all. 

There is, however, a severe decrease in performance worthy of note in \emph{gender\_identity}. We believe this might either be due to the split between train, validation, and test sets -- seeing as this class makes up a mere \emph{0.56\%} of Group D, and, as such, is easily affected by the random data split -- or due to some type of overlap of terms with the added \emph{gender} entries from the Single-Target datasets. The fact that there is no pattern in terms of Precision and Recall, in opposition to what we observed in the previous section, leads us to believe that the first option is the more likely answer.

We also see a noticeable improvement in \emph{gender}, \emph{religion}, and \emph{race}. The latter can, once more, be justified by the lower average value resulting from the anomalous result obtained in Multi-C, rather than any real improvement in the model's behaviour. 

The improvement observed in the other two classes, however, can be attributed directly to the addition of the Single-Target datasets which deal precisely with the target categories in question. The fact that the improvement in \emph{religion} is markedly lower than in \emph{gender} also supports this theory; \emph{religion} makes up \emph{1.43\%} of Group C's data compared to \emph{1.56\%} of Group D, while \emph{gender} goes from a modest \emph{1.72\%} in Group C to a respectable \emph{4.26\%} in Group D. Furthermore, we observed in Section \ref{group_b} that the Single-Target entries for ``Gender" behaved extremely well when added to the baseline datasets, which we attributed partly to the fact that all datasets in Group B were Twitter-based. CONAN's synthetic origin could be a contributing factor to the less marked improvement in performance.

\begin{figure}[h!t!]
    \centering
    \includegraphics[scale = 0.6]{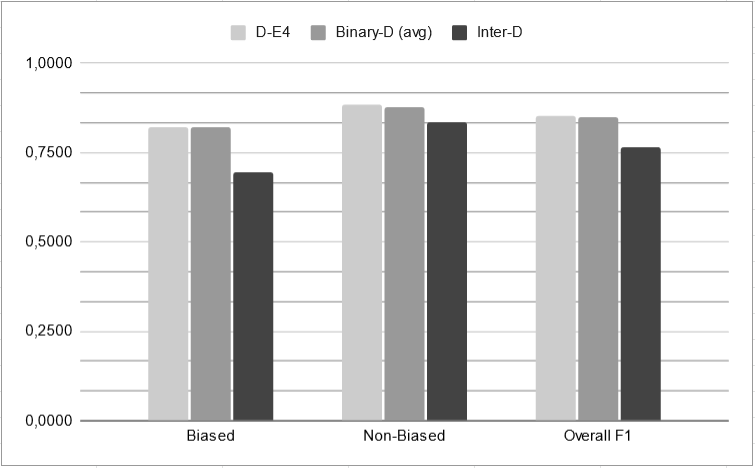}
    \caption{F1-scores of experiments Multi-D, Inter-D, as well as the average F1-scores of Binary-D}
    \label{fig:d_binary_compare}
\end{figure}

Shifting our attention to the Binary-Classification task, we can observe a pattern similar to the previous sections. The model trained with Group A data, used in the Inter-D experiment depicted in Table \ref{tab:inter_experiments}, does not perform nearly as well as the model trained with Group D data. Notably, it also performs noticeably worse in the Biased class, which can certainly be attributed to the synthetic and Multi-Target datasets featured in Group D, introducing not only new linguistic forms but also different ways to express and convey bias.

In the end, all our resources together do not perform better than our baseline group. We do observe marked improvement in the classes to which we added new entries when compared to those same classes in Group C, which suggests that the lower performance score might be due to the low number of entries for the several categories rather than the model's inherent difficulty in dealing with the different kinds of biases and categories. Additionally, while the synthetic datasets do not perform as well when the baseline data is Twitter-based, their addition to the training data is markedly necessary if we want a model that can properly identify them, as shown in the comparison between Inter-D and Binary-D.


\section{Conclusion}\label{conclusions}
Bias in NLP is a recent field of study, with plenty of works being published in recent years. We are discovering that there are many ways in which human biases can, and do, infiltrate our programs and algorithms. One of these ways is through biased training data, which teaches models how to replicate those very same biases.

In our work, we sought to use publicly available resources to train a classifier in the task of Bias Detection and Classification. The aim of our work was to discover if (or how) pre-existing resources could be used together to train a classifier in this task.

We find that while models can learn to identify Bias for a certain target category when trained when unspecified Bias/Hate Speech Detection datasets and a smaller dataset for that very same target category (Single-Target Classification), they do not perform well if one follows this system with many target categories and smaller datasets of varying sizes. However, models trained in this way still appear to be better at identifying Bias in synthetic text, or in more nuanced forms, than datasets trained only on generalized datasets and Twitter-based data, which implies that the model learns additional information that allows it to perform better in select contexts.

These conclusions emphasize the disproportionate attention given to certain targets of bias, which means that there are not enough resources available to train models to identify other types of biases. This is made worse by the reliance on non-persistent data, which leads to dataset degradation and further sabotages whatever available resources exist.

These conclusions also emphasize the need for clarity and diversity in further research in this field. It is paramount to diversify the focus of research, especially in an age in which social biases continue to grow in social importance. Technological advances must keep pace with societal ones, and that goal cannot be achieved if we remain stagnant and do not pay heed to recurring mistakes.

\section*{Acknowledgements}
This work was supported by Fundação para a Ciência e a Tecnologia (FCT), through Portuguese national funds Ref. UIDB/50021/2020, and Agência Nacional de Inovação (ANI), through the project C645008882-00000055 (Center for Responsible AI) funded by Recovery and Resilience Plan (RRP) and Next Generation EU European Funds and through the project CMU-PT MAIA Ref. 045909.

\bibliographystyle{unsrt}
\bibliography{references}

\begin{thebibliography}{10}

\bibitem{founta2018large}
Antigoni Founta, Constantinos Djouvas, Despoina Chatzakou, Ilias Leontiadis, Jeremy Blackburn, Gianluca Stringhini, Athena Vakali, Michael Sirivianos, and Nicolas Kourtellis.
\newblock Large scale crowdsourcing and characterization of twitter abusive behavior.
\newblock In {\em Proceedings of the International AAAI Conference on Web and Social Media}, volume~12, 2018.

\bibitem{field-etal-2021-survey}
Anjalie Field, Su~Lin Blodgett, Zeerak Waseem, and Yulia Tsvetkov.
\newblock A survey of race, racism, and anti-racism in {NLP}.
\newblock In {\em Proceedings of the 59th Annual Meeting of the Association for Computational Linguistics and the 11th International Joint Conference on Natural Language Processing (Volume 1: Long Papers)}, pages 1905--1925, Online, August 2021. Association for Computational Linguistics.

\bibitem{crenshaw2018demarginalizing}
Kimberl{\'e} Crenshaw.
\newblock {\em Demarginalizing the intersection of race and sex: A Black feminist critique of antidiscrimination doctrine, feminist theory, and antiracist politics [1989]}.
\newblock Routledge, 2018.

\bibitem{bolukbasi2016man}
Tolga Bolukbasi, Kai-Wei Chang, James~Y Zou, Venkatesh Saligrama, and Adam~T Kalai.
\newblock Man is to computer programmer as woman is to homemaker? debiasing word embeddings.
\newblock {\em Advances in neural information processing systems}, 29:4349--4357, 2016.

\bibitem{basta2019evaluating}
Christine Basta, Marta~R Costa-juss{\`a}, and Noe Casas.
\newblock Evaluating the underlying gender bias in contextualized word embeddings.
\newblock In {\em Proceedings of the First Workshop on Gender Bias in Natural Language Processing}, pages 33--39, 2019.

\bibitem{10.1145/3461702.3462536}
Wei Guo and Aylin Caliskan.
\newblock {\em Detecting Emergent Intersectional Biases: Contextualized Word Embeddings Contain a Distribution of Human-like Biases}, page 122–133.
\newblock Association for Computing Machinery, New York, NY, USA, 2021.

\bibitem{jiang-fellbaum-2020-interdependencies}
May Jiang and Christiane Fellbaum.
\newblock Interdependencies of gender and race in contextualized word embeddings.
\newblock In {\em Proceedings of the Second Workshop on Gender Bias in Natural Language Processing}, pages 17--25, Barcelona, Spain (Online), December 2020. Association for Computational Linguistics.

\bibitem{kaneko2019gender}
Masahiro Kaneko and Danushka Bollegala.
\newblock Gender-preserving debiasing for pre-trained word embeddings.
\newblock In {\em Proceedings of the 57th Annual Meeting of the Association for Computational Linguistics}, pages 1641--1650, 2019.

\bibitem{tan2019assessing}
Yi~Chern Tan and L~Elisa Celis.
\newblock Assessing social and intersectional biases in contextualized word representations.
\newblock {\em arXiv preprint arXiv:1911.01485}, 2019.

\bibitem{sharifirad2019learning}
Sima Sharifirad, Alon Jacovi, Israel Bar~Ilan Univesity, and Stan Matwin.
\newblock Learning and understanding different categories of sexism using convolutional neural network’s filters.
\newblock In {\em Proceedings of the 2019 Workshop on Widening NLP}, pages 21--23, 2019.

\bibitem{nadeem-etal-2021-stereoset}
Moin Nadeem, Anna Bethke, and Siva Reddy.
\newblock {S}tereo{S}et: Measuring stereotypical bias in pretrained language models.
\newblock In {\em Proceedings of the 59th Annual Meeting of the Association for Computational Linguistics and the 11th International Joint Conference on Natural Language Processing (Volume 1: Long Papers)}, pages 5356--5371, Online, August 2021. Association for Computational Linguistics.

\bibitem{parikh2019multi}
Pulkit Parikh, Harika Abburi, Pinkesh Badjatiya, Radhika Krishnan, Niyati Chhaya, Manish Gupta, and Vasudeva Varma.
\newblock Multi-label categorization of accounts of sexism using a neural framework.
\newblock In {\em Proceedings of the 2019 Conference on Empirical Methods in Natural Language Processing and the 9th International Joint Conference on Natural Language Processing (EMNLP-IJCNLP)}, pages 1642--1652, 2019.

\bibitem{liu-etal-2020-mitigating}
Haochen Liu, Wentao Wang, Yiqi Wang, Hui Liu, Zitao Liu, and Jiliang Tang.
\newblock Mitigating gender bias for neural dialogue generation with adversarial learning.
\newblock In {\em Proceedings of the 2020 Conference on Empirical Methods in Natural Language Processing (EMNLP)}, pages 893--903, Online, November 2020. Association for Computational Linguistics.

\bibitem{zhang2018mitigating}
Brian~Hu Zhang, Blake Lemoine, and Margaret Mitchell.
\newblock Mitigating unwanted biases with adversarial learning.
\newblock In {\em Proceedings of the 2018 AAAI/ACM Conference on AI, Ethics, and Society}, pages 335--340, 2018.

\bibitem{zhao2018gender}
Jieyu Zhao, Tianlu Wang, Mark Yatskar, Vicente Ordonez, and Kai-Wei Chang.
\newblock Gender bias in coreference resolution: Evaluation and debiasing methods.
\newblock In {\em Proceedings of the 2018 Conference of the North American Chapter of the Association for Computational Linguistics: Human Language Technologies, Volume 2 (Short Papers)}, pages 15--20, 2018.

\bibitem{kiritchenko-mohammad-2018-examining}
Svetlana Kiritchenko and Saif Mohammad.
\newblock Examining gender and race bias in two hundred sentiment analysis systems.
\newblock In {\em Proceedings of the Seventh Joint Conference on Lexical and Computational Semantics}, pages 43--53, New Orleans, Louisiana, June 2018. Association for Computational Linguistics.

\bibitem{dinan-etal-2020-queens}
Emily Dinan, Angela Fan, Adina Williams, Jack Urbanek, Douwe Kiela, and Jason Weston.
\newblock Queens are powerful too: Mitigating gender bias in dialogue generation.
\newblock In {\em Proceedings of the 2020 Conference on Empirical Methods in Natural Language Processing (EMNLP)}, pages 8173--8188, Online, November 2020. Association for Computational Linguistics.

\bibitem{garimella2019women}
Aparna Garimella, Carmen Banea, Dirk Hovy, and Rada Mihalcea.
\newblock Women’s syntactic resilience and men’s grammatical luck: Gender-bias in part-of-speech tagging and dependency parsing.
\newblock In {\em Proceedings of the 57th Annual Meeting of the Association for Computational Linguistics}, pages 3493--3498, 2019.

\bibitem{luccioni-viviano-2021-whats}
Alexandra Luccioni and Joseph Viviano.
\newblock What{'}s in the box? an analysis of undesirable content in the {C}ommon {C}rawl corpus.
\newblock In {\em Proceedings of the 59th Annual Meeting of the Association for Computational Linguistics and the 11th International Joint Conference on Natural Language Processing (Volume 2: Short Papers)}, pages 182--189, Online, August 2021. Association for Computational Linguistics.

\bibitem{wolf2017we}
Marty~J Wolf, Keith~W Miller, and Frances~S Grodzinsky.
\newblock Why we should have seen that coming: comments on microsoft’s tay “experiment,” and wider implications.
\newblock {\em The ORBIT Journal}, 1(2):1--12, 2017.

\bibitem{borji2023categorical}
Ali Borji.
\newblock A categorical archive of chatgpt failures.
\newblock {\em arXiv e-prints}, pages arXiv--2302, 2023.

\bibitem{baker-gillis-2021-sexism}
Noa Baker~Gillis.
\newblock Sexism in the judiciary: The importance of bias definition in {NLP} and in our courts.
\newblock In {\em Proceedings of the 3rd Workshop on Gender Bias in Natural Language Processing}, pages 45--54, Online, August 2021. Association for Computational Linguistics.

\bibitem{park2021multilingual}
Chan~Young Park, Xinru Yan, Anjalie Field, and Yulia Tsvetkov.
\newblock Multilingual contextual affective analysis of lgbt people portrayals in wikipedia.
\newblock In {\em Proceedings of the International AAAI Conference on Web and Social Media}, volume~15, pages 479--490, 2021.

\bibitem{touileb2020gender}
Samia Touileb, Lilja {\O}vrelid, and Erik Velldal.
\newblock Gender and sentiment, critics and authors: a dataset of norwegian book reviews.
\newblock In {\em Proceedings of the Second Workshop on Gender Bias in Natural Language Processing}, pages 125--138, 2020.

\bibitem{badjatiya2017deep}
Pinkesh Badjatiya, Shashank Gupta, Manish Gupta, and Vasudeva Varma.
\newblock Deep learning for hate speech detection in tweets.
\newblock In {\em Proceedings of the 26th international conference on World Wide Web companion}, pages 759--760, 2017.

\bibitem{davidson2017automated}
Thomas Davidson, Dana Warmsley, Michael Macy, and Ingmar Weber.
\newblock Automated hate speech detection and the problem of offensive language.
\newblock In {\em Proceedings of the International AAAI Conference on Web and Social Media}, volume~11, 2017.

\bibitem{djuric2015hate}
Nemanja Djuric, Jing Zhou, Robin Morris, Mihajlo Grbovic, Vladan Radosavljevic, and Narayan Bhamidipati.
\newblock Hate speech detection with comment embeddings.
\newblock In {\em Proceedings of the 24th international conference on world wide web}, pages 29--30, 2015.

\bibitem{golbeck2017large}
Jennifer Golbeck, Zahra Ashktorab, Rashad~O Banjo, Alexandra Berlinger, Siddharth Bhagwan, Cody Buntain, Paul Cheakalos, Alicia~A Geller, Rajesh~Kumar Gnanasekaran, Raja~Rajan Gunasekaran, et~al.
\newblock A large labeled corpus for online harassment research.
\newblock In {\em Proceedings of the 2017 ACM on web science conference}, pages 229--233, 2017.

\bibitem{fersini2018overview}
Elisabetta Fersini, Paolo Rosso, and Maria Anzovino.
\newblock Overview of the task on automatic misogyny identification at ibereval 2018.
\newblock {\em IberEval@ SEPLN}, 2150:214--228, 2018.

\bibitem{jha2017does}
Akshita Jha and Radhika Mamidi.
\newblock When does a compliment become sexist? analysis and classification of ambivalent sexism using twitter data.
\newblock In {\em Proceedings of the second workshop on NLP and computational social science}, pages 7--16, 2017.

\bibitem{suvarna2020notawhore}
Ashima Suvarna and Grusha Bhalla.
\newblock \# notawhore! a computational linguistic perspective of rape culture and victimization on social media.
\newblock In {\em Proceedings of the 58th Annual Meeting of the Association for Computational Linguistics: Student Research Workshop}, pages 328--335, 2020.

\bibitem{chung-etal-2019-conan}
Yi-Ling Chung, Elizaveta Kuzmenko, Serra~Sinem Tekiroglu, and Marco Guerini.
\newblock {CONAN} - {CO}unter {NA}rratives through nichesourcing: a multilingual dataset of responses to fight online hate speech.
\newblock In {\em Proceedings of the 57th Annual Meeting of the Association for Computational Linguistics}, pages 2819--2829, Florence, Italy, July 2019. Association for Computational Linguistics.

\bibitem{barikeri-etal-2021-redditbias}
Soumya Barikeri, Anne Lauscher, Ivan Vuli{\'c}, and Goran Glava{\v{s}}.
\newblock {R}eddit{B}ias: A real-world resource for bias evaluation and debiasing of conversational language models.
\newblock In {\em Proceedings of the 59th Annual Meeting of the Association for Computational Linguistics and the 11th International Joint Conference on Natural Language Processing (Volume 1: Long Papers)}, pages 1941--1955, Online, August 2021. Association for Computational Linguistics.

\bibitem{vidgen-etal-2021-learning}
Bertie Vidgen, Tristan Thrush, Zeerak Waseem, and Douwe Kiela.
\newblock Learning from the worst: Dynamically generated datasets to improve online hate detection.
\newblock In {\em Proceedings of the 59th Annual Meeting of the Association for Computational Linguistics and the 11th International Joint Conference on Natural Language Processing (Volume 1: Long Papers)}, pages 1667--1682, Online, August 2021. Association for Computational Linguistics.

\bibitem{waseem2016hateful}
Zeerak Waseem and Dirk Hovy.
\newblock Hateful symbols or hateful people? predictive features for hate speech detection on twitter.
\newblock In {\em Proceedings of the NAACL student research workshop}, pages 88--93, 2016.

\bibitem{Poletto2021ResourcesAB}
Fabio Poletto, Valerio Basile, Manuela Sanguinetti, Cristina Bosco, and Viviana Patti.
\newblock Resources and benchmark corpora for hate speech detection: a systematic review.
\newblock {\em Lang. Resour. Evaluation}, 55:477--523, 2021.

\bibitem{vidgen-etal-2019-challenges}
Bertie Vidgen, Alex Harris, Dong Nguyen, Rebekah Tromble, Scott Hale, and Helen Margetts.
\newblock Challenges and frontiers in abusive content detection.
\newblock In {\em Proceedings of the Third Workshop on Abusive Language Online}, pages 80--93, Florence, Italy, August 2019. Association for Computational Linguistics.

\bibitem{fortuna-etal-2021-min}
Paula Fortuna, Vanessa Cortez, Miguel Sozinho~Ramalho, and Laura P{\'e}rez-Mayos.
\newblock {MIN}{\_}{PT}: An {E}uropean {P}ortuguese lexicon for minorities related terms.
\newblock In {\em Proceedings of the 5th Workshop on Online Abuse and Harms (WOAH 2021)}, pages 76--80, Online, August 2021. Association for Computational Linguistics.

\bibitem{zhou-etal-2019-examining}
Pei Zhou, Weijia Shi, Jieyu Zhao, Kuan-Hao Huang, Muhao Chen, Ryan Cotterell, and Kai-Wei Chang.
\newblock Examining gender bias in languages with grammatical gender.
\newblock In {\em Proceedings of the 2019 Conference on Empirical Methods in Natural Language Processing and the 9th International Joint Conference on Natural Language Processing (EMNLP-IJCNLP)}, pages 5276--5284, Hong Kong, China, November 2019. Association for Computational Linguistics.

\bibitem{sap-etal-2019-risk}
Maarten Sap, Dallas Card, Saadia Gabriel, Yejin Choi, and Noah~A. Smith.
\newblock The risk of racial bias in hate speech detection.
\newblock In {\em Proceedings of the 57th Annual Meeting of the Association for Computational Linguistics}, pages 1668--1678, Florence, Italy, July 2019. Association for Computational Linguistics.

\bibitem{excell-al-moubayed-2021-towards}
Elizabeth Excell and Noura Al~Moubayed.
\newblock Towards equal gender representation in the annotations of toxic language detection.
\newblock In {\em Proceedings of the 3rd Workshop on Gender Bias in Natural Language Processing}, pages 55--65, Online, August 2021. Association for Computational Linguistics.

\bibitem{10.1145/3411764.3445092}
Miriah Steiger, Timir~J Bharucha, Sukrit Venkatagiri, Martin~J. Riedl, and Matthew Lease.
\newblock The psychological well-being of content moderators: The emotional labor of commercial moderation and avenues for improving support.
\newblock In {\em Proceedings of the 2021 CHI Conference on Human Factors in Computing Systems}, CHI '21, New York, NY, USA, 2021. Association for Computing Machinery.

\bibitem{doi:10.1177/01634437221122226}
Amit Pinchevski.
\newblock Social media’s canaries: content moderators between digital labor and mediated trauma.
\newblock {\em Media, Culture \& Society}, 45(1):212--221, 2023.

\bibitem{nangia-etal-2020-crows}
Nikita Nangia, Clara Vania, Rasika Bhalerao, and Samuel~R. Bowman.
\newblock {C}row{S}-pairs: A challenge dataset for measuring social biases in masked language models.
\newblock In {\em Proceedings of the 2020 Conference on Empirical Methods in Natural Language Processing (EMNLP)}, pages 1953--1967, Online, November 2020. Association for Computational Linguistics.

\bibitem{ousidhoum-etal-2019-multilingual}
Nedjma Ousidhoum, Zizheng Lin, Hongming Zhang, Yangqiu Song, and Dit-Yan Yeung.
\newblock Multilingual and multi-aspect hate speech analysis.
\newblock In {\em Proceedings of the 2019 Conference on Empirical Methods in Natural Language Processing and the 9th International Joint Conference on Natural Language Processing (EMNLP-IJCNLP)}, pages 4675--4684, Hong Kong, China, November 2019. Association for Computational Linguistics.

\end{thebibliography}


\end{document}